\documentclass{article} 
\usepackage{iclr2027_conference,times}

\usepackage{amsmath,amsfonts,bm}

\def\eqref#1{equation~\ref{#1}}

\def\1{\bm{1}}

\DeclareMathAlphabet{\mathsfit}{\encodingdefault}{\sfdefault}{m}{sl}
\SetMathAlphabet{\mathsfit}{bold}{\encodingdefault}{\sfdefault}{bx}{n}

\usepackage{hyperref}
\usepackage{url}

\usepackage{graphicx}
\usepackage{booktabs}
\usepackage{tabularx}
\usepackage{array}
\usepackage{multirow}
\usepackage{makecell}
\usepackage{amsmath}
\usepackage{amssymb}
\usepackage{bm}
\usepackage{algorithm}
\usepackage{algorithmic}
\usepackage{adjustbox}
\usepackage{xcolor}
\usepackage{tcolorbox}
\tcbuselibrary{skins}
\usepackage{wrapfig}
\usepackage{graphicx}
\usepackage{caption}
\usepackage{booktabs}
\usepackage{tabularx}
\usepackage{array}
\usepackage{needspace}
\usepackage{enumitem}

\newcolumntype{C}{>{\centering\arraybackslash}X}

\providecommand{\introfigurewidth}{0.54\columnwidth}

\newcolumntype{L}{>{\raggedright\arraybackslash}X}
\newcolumntype{C}{>{\centering\arraybackslash}X}

\newcolumntype{C}{>{\centering\arraybackslash}X}

\title{CircuitGate: Logic-Consistent Circuit-Level Functional Modeling for And-Inverter Graphs}

\author{
\textbf{
Qifan Zhang$^{1,\uparrow}$,
Ruijie Li$^{2,\uparrow}$,
Fangzhou Zhang$^{1}$,
Qian Ma$^{1,*}$,
Hui Li$^{1}$,
Furui Zhan$^{1}$,
} \\
\textbf{
Yongpeng Wang$^{1}$,
Liying Hao$^{1}$,
Shikai Guo$^{1}$
} \\
\normalfont
$^{1}$Dalian Maritime University \\
$^{2}$The Hong Kong University of Science and Technology (Guangzhou) \\
$^{\uparrow}$Equal contribution \\
$^{*}$Corresponding author
}

\iclrfinalcopy

\begin{document}

\setlength{\emergencystretch}{1em}

\maketitle

\begin{abstract}
And-Inverter Graphs (AIGs) are fundamental representations for logic synthesis and verification in Electronic Design Automation (EDA).
As structured representations of complex digital systems, AIGs require models to capture functional dependencies beyond local structure and remain robust to functionality-preserving transformations.
In learning-based AIG representation, existing approaches are predominantly based on GNNs and rely on local gate-level message passing, limiting their ability to capture circuit-level functional context and making the learned representations sensitive to topology-specific patterns.
Therefore, we propose \textbf{CircuitGate}, a function-aware AIG representation learning framework that advances from gate-level semantics to circuit-level functional modeling.
CircuitGate explicitly encodes global primary-input (PI) support and models support-overlap-aware reconvergence between fanins, while incorporating logic-inspired Boolean constraints to encourage functionally consistent representations.
We evaluate CircuitGate on the large-scale ForgeEDA benchmark and further validate it on the EPFL and ITC'99 benchmarks.
Across equivalent-gate identification and signal-probability prediction tasks, CircuitGate consistently outperforms existing methods, achieving up to 21.7\% and 14.2\% reductions in MAE, respectively.
Under direct ForgeEDA-to-OpenABC transfer without fine-tuning, CircuitGate also achieves the best equivalent-gate identification performance, demonstrating strong cross-dataset generalization.
These results demonstrate the effectiveness of modeling circuit-level functional dependencies beyond local topology.
Source code is available at \url{https://anonymous.4open.science/r/CircuitGate/}.
\end{abstract}

\section{Introduction}
And-Inverter Graphs (AIGs) are fundamental intermediate representations in Electronic Design Automation (EDA), supporting critical tasks such as logic synthesis, formal verification, and optimization~\cite{wang2009electronic,hassoun2001logic}.
By representing Boolean networks using two-input AND gates and complemented edges, AIGs provide a compact and widely adopted abstraction for EDA workflows, including technology mapping~\cite{mishchenko2007abc} and equivalence checking~\cite{goldberg2001using,kern1999formal}.
As circuit scale and logic complexity continue to grow, learning informative representations of AIGs has become important for enabling effective learning-based EDA applications.
However, a useful AIG representation should capture more than the local structural patterns of an implementation; it should reflect the underlying circuit functionality that remains stable across structurally different but functionally equivalent implementations. This requires modeling circuit-level functional dependencies beyond local topology while maintaining consistent semantics under functionality-preserving synthesis transformations.

\begin{wrapfigure}{r}{\introfigurewidth} \centering \includegraphics[width=\linewidth]{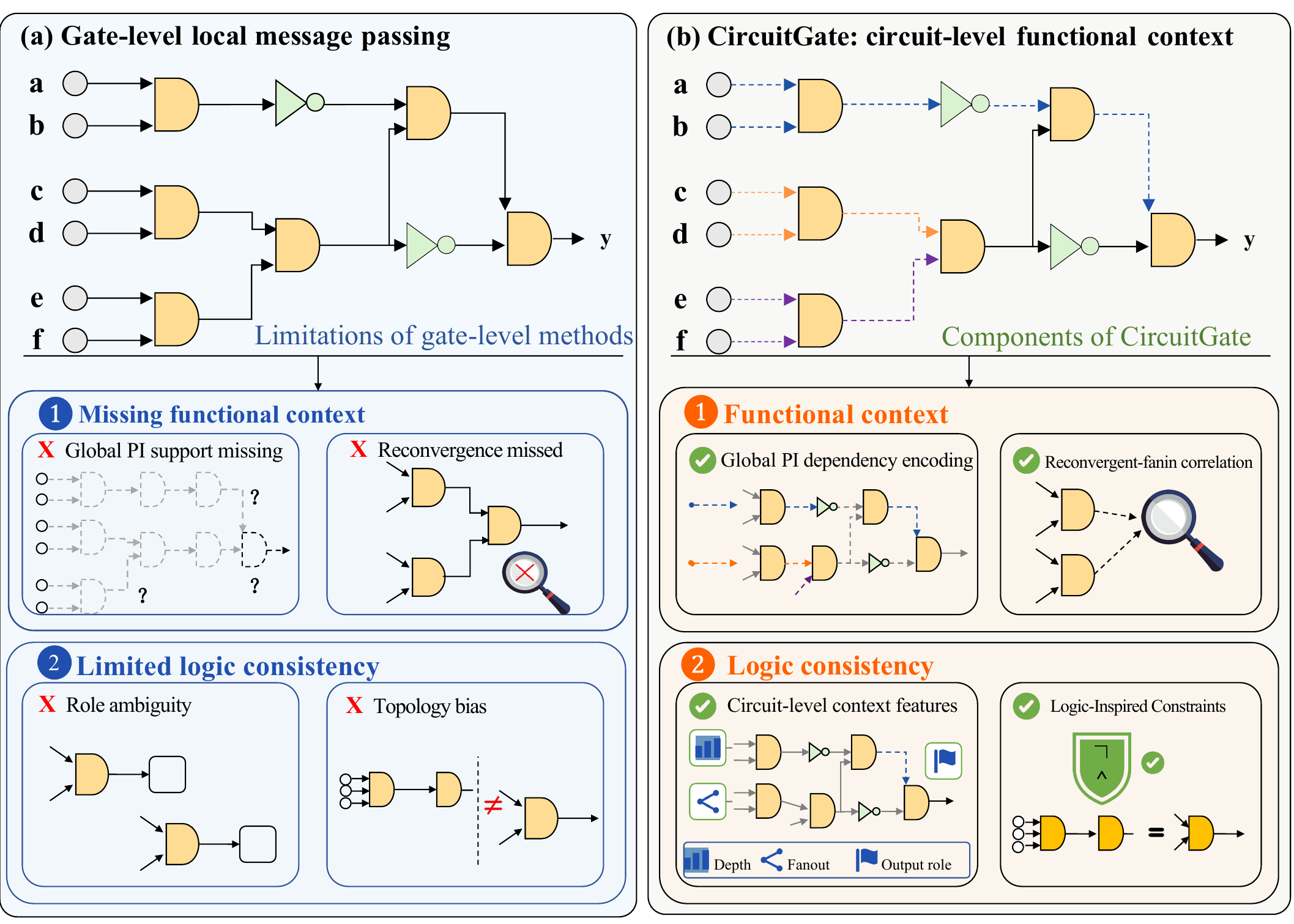} \captionsetup{ font=small, justification=raggedright, singlelinecheck=false, skip=3pt } \caption{Motivating comparison between prior gate-level AIG representation methods and CircuitGate. Existing approaches mainly rely on local structural message passing, resulting in incomplete functional context and limited logic consistency, while CircuitGate explicitly models circuit-level functional dependencies and improves representation consistency.} \label{fig:intro-comparison} \end{wrapfigure}

Recent advances in Graph Neural Networks (GNNs) have enabled learning-based AIG representation methods.
DeepGate~\cite{li2022deepgate} pioneered GNN-based AIG representation learning by introducing signal probability supervision.
Subsequent works, including DeepGate2~\cite{shi2023deepgate2}, DeepGate3~\cite{shi2024deepgate3}, and DeepGate4~\cite{zheng2025deepgate4}, further improved supervision strategies, scalability, and industrial applicability.
Other approaches, including FGNN~\cite{wang2022functionality}, FuncGNN~\cite{zhao2026funcgnn}, PolarGate~\cite{liu2024polargate}, HOGA~\cite{deng2024less}, and WideGate~\cite{liu2025widegate}, explored complementary directions such as functionality-aware propagation, hybrid aggregation, polarity modeling, and graph attention mechanisms.
Despite these advances, as illustrated in Figure~\ref{fig:intro-comparison}, existing AIG representation methods still propagate information primarily through local connectivity patterns, limiting their ability to capture complete circuit-level functional dependencies and maintain consistent representations across different circuit implementations.
These limitations manifest in two key aspects.

First, existing methods provide an incomplete characterization of circuit-level functional dependencies. 
Local message passing is effective for modeling Boolean operations among neighboring gates and capturing short-range structural interactions. 
However, the functionality of a circuit node is ultimately determined by the complete set of primary inputs (PIs) that influence its output. 
Such global PI support provides essential dependency information beyond local neighborhoods but is difficult to preserve through finite-hop message propagation, especially in large circuits with deep logic hierarchies. 
Moreover, industrial circuits contain extensive reconvergent structures, where multiple fanin branches with shared PI dependencies merge at downstream nodes. 
These fanins are therefore functionally correlated rather than independent structural neighbors. 
Existing message passing mechanisms process these fanins independently and fail to explicitly model their shared-support relationships.
Consequently, current methods lack a unified mechanism for capturing global PI dependencies and reconvergent-fanin correlations, resulting in incomplete circuit-level functional representations.

Second, existing methods exhibit limited consistency under functionality-preserving synthesis transformations. 
Modern synthesis procedures, such as \texttt{resyn2}~\cite{mishchenko2007abc} and \texttt{dc2}, can substantially modify AIG topology, including node structures, logic depth, and connectivity patterns, while preserving Boolean functionality. 
However, topology-driven representation models may encode implementation-specific structural patterns rather than intrinsic functional properties.
As a result, functionally equivalent circuits generated by different synthesis processes may produce inconsistent representations.
This limitation highlights a broader challenge in representation learning for structured engineering systems: effective representations should capture stable functional characteristics rather than rely primarily on transformation-dependent topology.

These observations lead to a central challenge:
\begin{tcolorbox}[enhanced, width=0.95\linewidth, center,
    colback=white, colframe=gray!60!black, boxrule=1pt,
    arc=2mm, drop fuzzy shadow]
    \centering \itshape
    How can we learn AIG representations that capture complete circuit-level functional dependencies beyond local topology while remaining consistent under functionality-preserving synthesis transformations?
\end{tcolorbox}

To address this challenge, we propose \textbf{CircuitGate}, a function-aware representation learning framework for AIGs. 
On the one hand, CircuitGate explicitly models circuit-level functional dependencies by incorporating global PI-support information and support-overlap-aware reconvergent-fanin relationships, enabling each node to capture functional context beyond its local structural neighborhood. 
On the other hand, CircuitGate augments topology-driven message passing with complementary functional contexts and logic-inspired Boolean constraints, encouraging functionally equivalent circuits to exhibit more consistent representations despite synthesis-induced structural variations. 
Together, these designs advance AIG representation learning from local gate-level semantic modeling toward logic-consistent circuit-level functional representation learning.
Extensive experiments primarily on the large-scale ForgeEDA benchmark, with additional validation on EPFL and ITC'99 and zero-shot ForgeEDA-to-OpenABC transfer, demonstrate the effectiveness and superiority of CircuitGate across multiple AIG representation learning tasks and circuit distributions.

The main contributions of this work are:
\begin{itemize}[leftmargin=*]
    \item We provide a systematic analysis of existing AIG representation learning methods and identify a fundamental gap between gate-level semantic modeling and circuit-level functional representation. 
    Specifically, current approaches lack effective mechanisms for capturing global functional dependencies and are vulnerable to topology-specific biases introduced by functionality-preserving transformations, motivating the design of CircuitGate.

    \item We propose CircuitGate, a unified framework that advances AIG representation learning from local topology-driven modeling toward circuit-level functional dependency modeling.
    CircuitGate jointly captures global PI dependencies and support-overlap-aware reconvergence through circuit-level context modeling, global dependency encoding, and reconvergent-fanin dependency learning, while incorporating Boolean constraints to encourage functionally consistent representations.

    \item We primarily evaluate CircuitGate on the large-scale ForgeEDA industrial EDA
benchmark, comprising 83,155 circuit instances derived from 4,450 AIGs across
1,189 open-source designs, and further evaluate it on EPFL and ITC'99.
CircuitGate reduces MAE over the second-best methods by 21.7\% and 13.8\% on
ForgeEDA for equivalent-gate identification and signal-probability prediction,
respectively. On EPFL, the corresponding reductions are 19.5\% and 14.2\%,
and on ITC'99, 13.4\% and 8.4\%.
Zero-shot ForgeEDA-to-OpenABC transfer further demonstrates cross-dataset
generalization, while ablation studies verify the contributions of individual
components.
\end{itemize}

\begin{figure*}[t]
  \centering
  \scalebox{1}[1.12]{%
    \includegraphics[width=\linewidth]{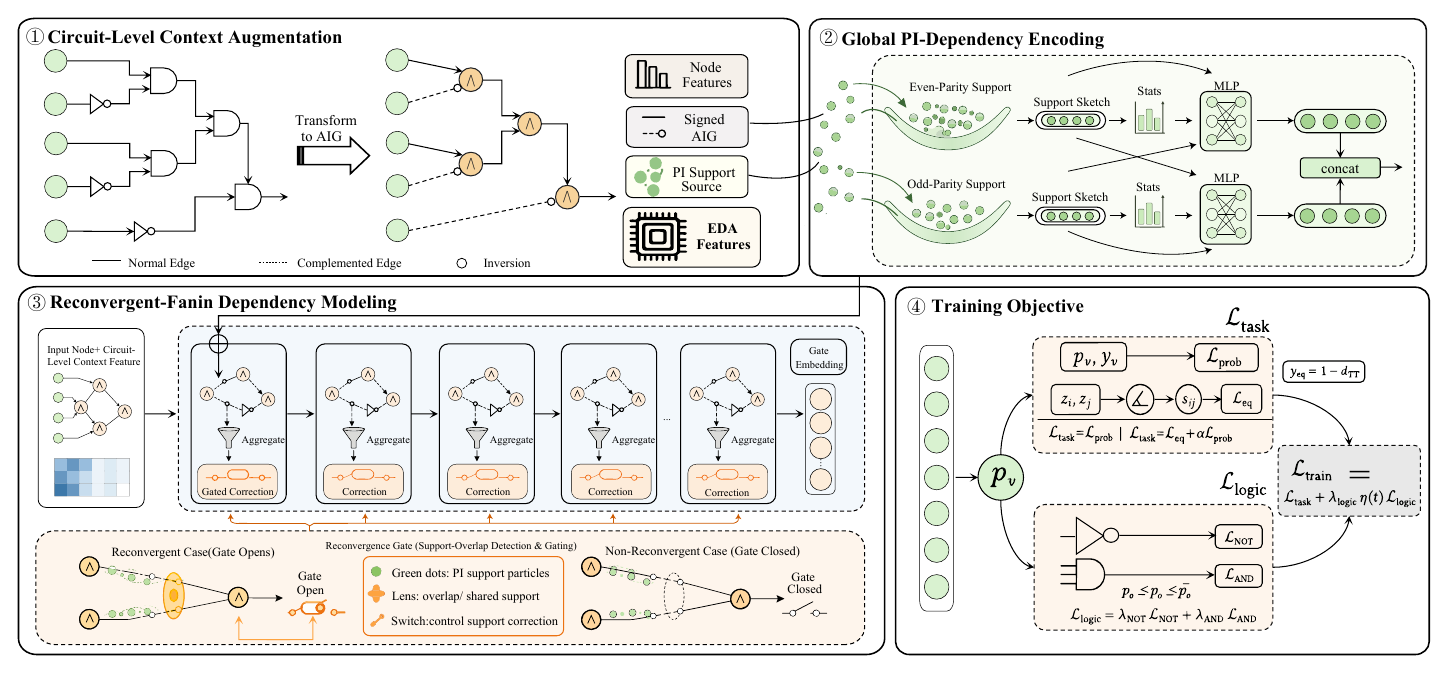}
  }
  \captionsetup{
    font=small,
    skip=3pt
  }
  \caption{Overview of the CircuitGate framework.}
  \label{fig:overview}
  \vspace{-13pt}
\end{figure*}
\section{Related Work}
\label{sec:related-work}

\subsection{Functional Dependency Modeling}

Traditional support-based AIG analysis explicitly tracks primary-input (PI)
dependencies through structural support information~\cite{mishchenko2007abc},
making circuit-level dependencies directly available for logic reasoning.
Modern GNN-based methods recast dependency modeling as scalable implicit
message propagation.
DeepGate~\cite{li2022deepgate} pioneered GNN-based gate representation
learning with signal-probability supervision, followed by
DeepGate2~\cite{shi2023deepgate2}, DeepGate3~\cite{shi2024deepgate3}, and
DeepGate4~\cite{zheng2025deepgate4}, which progressively advance functional
modeling and scalability.
Recent methods further explore functionality-aware and circuit-specific
modeling, including GAMORA~\cite{wu2023gamora}, FGNN~\cite{wang2022functionality},
FuncGNN~\cite{zhao2026funcgnn}, AIGer~\cite{sun2025aiger},
WideGate~\cite{liu2025widegate}, PolarGate~\cite{liu2024polargate}, and
HOGA~\cite{deng2024less}.

This transition improves scalability but makes circuit-level dependency
modeling increasingly reliant on local propagation.
In standard message-passing GNNs, $k$ layers aggregate information over at
most $k$-hop neighborhoods, making global PI dependencies difficult to
preserve across deep logic hierarchies.
Moreover, scalable GNN-based methods still lack explicit reconvergence
modeling within representation updates: fanins with shared PI support are
typically aggregated as local neighbors, leaving their shared-input
dependencies only indirectly reflected in learned representations.
CircuitGate addresses these limitations through global PI-support encoding
and support-overlap-conditioned reconvergence modeling.

\subsection{Synthesis Robustness}

Logic synthesis exposes a fundamental tension in circuit representation
learning: topology can change dramatically while Boolean functionality
remains unchanged.
MILS~\cite{zhao2025mils}, CHOP~\cite{fu2026chop}, and
ORL-LO~\cite{dong2026orl} explore learning-based synthesis prediction and
optimization, while DeepCell~\cite{shi2025deepcell} and
GenEDA~\cite{xu2025geneda} study multiview and cross-modal circuit
representation learning.
TopoRTL~\cite{zhao2026topoRTL} and IRGNN~\cite{guo2025irgnn} further extend
circuit learning across abstraction levels and application domains.

Despite this progress, representation consistency under functionality-preserving
topology transformations remains a distinct challenge.
CircuitGate targets this setting through circuit-level context, global
PI-support encoding, reconvergence-aware modeling, and logic-inspired
regularization to reduce reliance on topology-specific patterns.
\section{Method}

\subsection{Overview}

Given an AIG represented as a signed directed acyclic graph
$\mathcal{G}=(\mathcal{V},\mathcal{E})$, where each edge carries an inversion label
$r_e\in\{0,1\}$, CircuitGate aims to learn node representations that capture
circuit-level functional dependencies beyond local gate-level semantics.
Different from topology-driven representations, CircuitGate seeks to preserve
global input dependencies and reconvergent-fanin correlations while reducing
reliance on synthesis-induced structural patterns.

The key idea is to augment local message passing with two complementary
sources of functional context:
(i) global PI dependencies that characterize the input influence of each node,
and (ii) local reconvergent-fanin correlations induced by shared PI supports.
As illustrated in Figure~\ref{fig:overview}, CircuitGate consists of three
architectural components:
\textbf{(1) Circuit-Level Context Augmentation}, which incorporates EDA-native
circuit-role information, including logic levels, fanout, output status, and
PI-support coverage, to enrich gate-level representations with circuit-level
contextual priors;
\textbf{(2) Global PI-Dependency Encoding}, which explicitly encodes
polarity-aware PI dependencies and compresses variable-size support sets into
fixed-width representations, enabling the model to capture global functional
dependencies beyond finite-hop message passing; and
\textbf{(3) Reconvergent-Fanin Dependency Modeling}, which identifies
shared-support relationships among reconvergent fanins and uses
support-overlap-aware gating to adaptively modulate the contribution of
PI-dependency context during node updates.
Beyond these architectural components, logic-inspired constraints are
incorporated into the training objective to regularize local Boolean
semantics and reduce topology-specific biases.
Together, these designs enable CircuitGate to learn circuit-level
functional representations with logic-inspired regularization.

\subsection{Circuit-Level Context Augmentation}
\label{sec:feat}

Logic synthesis can alter local wiring patterns while preserving circuit
functionality, making gate type insufficient to characterize a node's
circuit-level role.
To address this limitation, the Circuit-Level Context Augmentation component
augments each node's gate-type representation with EDA-native attributes
describing circuit position, fanout, output status, and PI-support coverage.

Specifically, the EDA-native context vector and the initial node
representation are defined as
\begin{equation}
\begin{aligned}
\mathbf{x}_v^{\mathrm{eda}}
&=
\big[
\bar{\ell}_v^{\mathrm{f}},
\bar{\ell}_v^{\mathrm{b}},
\bar{d}_v,
\delta_v^{\mathrm{out}},
\rho_v
\big],
\qquad
\mathbf{x}_v
&=
\big[
\mathbf{o}_v
\mathbin{\Vert}
\mathbf{x}_v^{\mathrm{eda}}
\big],
\end{aligned}
\label{eq:eda-features}
\end{equation}
where $\mathbf{o}_v$ is the gate-type one-hot vector and $\mathbin{\Vert}$
denotes concatenation.
For $k\in\{\mathrm{f},\mathrm{b}\}$, $\ell_v^k$ is the corresponding logic
level and
$\bar{\ell}_v^k=
\ell_v^k/\max\{1,\max_{u\in\mathcal{V}}\ell_u^k\}$.

$\bar{d}_v=\log(1+d_{\mathrm{out}}(v))$ denotes the transformed fanout, and
$\delta_v^{\mathrm{out}}
=\mathbf{1}_{\{d_{\mathrm{out}}(v)=0\}}$
indicates whether $v$ is a sink node.
$\rho_v$ represents PI-support coverage, defined as the larger fraction of
PIs in the even- and odd-inversion support sets
$\mathcal{S}_0(v)$ and $\mathcal{S}_1(v)$, with $|\mathcal{P}|$ as the
normalization factor.

These features distinguish same-type nodes by circuit role, helping the model
separate structurally different candidates before message passing begins.

\subsection{Global PI-Dependency Encoding}
\label{sec:support}

Nodes with similar local neighborhoods may depend on different primary inputs
(PIs), causing finite-depth message passing to conflate distinct global
dependencies.
The Global PI-Dependency Encoding component therefore summarizes the
structural PI dependencies of each node in fixed-width representations.
Because complemented edges alter the polarity with which a PI reaches a
downstream node, these dependencies are further distinguished by the parity
of inversions along PI-to-node paths.

Let $\mathcal{P}\subseteq\mathcal{V}$ denote the set of PIs and define
$P_0=\max\{|\mathcal{P}|,1\}$.
For each node $v$, we construct two parity-indexed structural PI-support sets,
where $q\in\{0,1\}$ denotes the inversion parity of a PI-to-$v$ path.
Let $\mathcal{N}_r(v)$ denote the fanins of $v$ connected through edges with
inversion label $r\in\{0,1\}$.
For each non-PI node $v$, the support sets are propagated in one
topological pass as
\begin{align}
\mathcal{S}_q(v)
&=
\bigcup_{u\in\mathcal{N}_0(v)}
\mathcal{S}_q(u)
\cup
\bigcup_{u\in\mathcal{N}_1(v)}
\mathcal{S}_{q\oplus 1}(u),
\label{eq:support-propagation}
\end{align}
where $\oplus$ denotes modulo-two addition.
Each PI $p$ is initialized by $\mathcal{S}_0(p)=\{p\}$ and
$\mathcal{S}_1(p)=\emptyset$.

These sets provide a path-based characterization of parity-aware
circuit-level PI dependencies, while serving as a structural approximation
of exact Boolean functional support.
In particular, a PI may belong to both $\mathcal{S}_0(v)$ and
$\mathcal{S}_1(v)$ if it reaches $v$ through paths with both inversion
parities.

To compress the variable-size supports, each PI is assigned a graph-local
index and mapped to a random token
$\mathbf{t}_p\sim\mathcal{N}(\mathbf{0},\mathbf{I}/d_s)$ using a fixed seed.
The tokens remain unchanged throughout training and inference.
Each parity-indexed support set is then summarized by the sketch
\begin{align}
\mathbf{s}_q(v)
&=
\frac{
\sum_{p\in\mathcal{S}_q(v)}
\mathbf{t}_p
}{
\sqrt{\max\{|\mathcal{S}_q(v)|,1\}}
},
\label{eq:support-sketch}
\end{align}
where $\mathbf{s}_q(v)\in\mathbb{R}^{d_s}$ and an empty support is mapped to
the zero vector.
The square-root normalization stabilizes the sketch variance across support
sizes, while explicit support statistics retain scale information.
The resulting sketch provides a compact randomized summary of PI-support
identity.
The sketches are further fused with support-size and PI-coverage statistics:
\begin{equation}
\begin{aligned}
\phi_q(v)
&=
\Big[
\log\!\big(1+|\mathcal{S}_q(v)|\big),
\log\!\big(1+|\mathcal{S}_{q\oplus1}(v)|\big),
|\mathcal{S}_q(v)|/P_0,
|\mathcal{S}_{q\oplus1}(v)|/P_0
\Big],\\
\mathbf{e}_q(v)
&=
\mathrm{MLP}_q\!\Big(
[
\mathbf{s}_q(v)
\mathbin{\Vert}
\mathbf{s}_{q\oplus1}(v)
\mathbin{\Vert}
\phi_q(v)
]
\Big).
\end{aligned}
\label{eq:support-encoding}
\end{equation}
where
$\operatorname{MLP}_q:
\mathbb{R}^{2d_s+4}\rightarrow\mathbb{R}^{d_h}$
is a learned state-specific multilayer perceptron and
$\mathbf{e}_q(v)\in\mathbb{R}^{d_h}$ is the encoded PI-dependency
representation for state $q$.
The two parity states are combined as
\begin{equation}
\mathbf{e}_v
=
[\mathbf{e}_0(v)\mathbin{\Vert}\mathbf{e}_1(v)]
\in\mathbb{R}^{2d_h}.
\end{equation}

The encoded PI dependencies are introduced after the first propagation layer
and concatenated into every subsequent state update while remaining unchanged
across layers.
This provides each node with a circuit-level prior for modeling functional
dependencies beyond finite-hop neighborhoods, improving discrimination
between locally similar nodes with different global dependencies.

\subsection{Reconvergent-Fanin Dependency Modeling}
\label{sec:reconv}

PI-support overlap provides structural evidence of shared upstream dependency
between reconvergent fanins.
CircuitGate characterizes reconvergence through the overlap between the PI
supports of two fanins and uses this information to condition local message
propagation.
The resulting mechanism allows global PI dependencies to adaptively influence
node updates while preserving the original AIG connectivity.

For a two-input AND node $v$ with fanins $a$ and $b$, let
$\mathcal{S}_x=\mathcal{S}_0(x)\cup\mathcal{S}_1(x)$ for
$x\in\{a,b\}$.
Its six-dimensional reconvergence descriptor is defined by
\begin{equation}
\begin{aligned}
n_v
&=
|\mathcal{S}_a\cap\mathcal{S}_b|,
\qquad
c_v^{\mathrm{s}}
=
\textstyle\sum_{q\in\{0,1\}}
|\mathcal{S}_q(a)\cap\mathcal{S}_q(b)|,
\qquad
c_v^{\mathrm{o}}
=
\textstyle\sum_{q\in\{0,1\}}
|\mathcal{S}_q(a)\cap\mathcal{S}_{q\oplus1}(b)|,\\[2pt]
\mathbf{r}_v
&=
\Bigg[
\mathbf{1}_{\{n_v>0\}},
\frac{n_v}{\max\{|\mathcal{S}_a\cup\mathcal{S}_b|,1\}},
\frac{n_v}{\max\{\min(|\mathcal{S}_a|,|\mathcal{S}_b|),1\}},
\\[-1pt]
&\qquad
\log(1+n_v),
\frac{|\ell^{\mathrm{f}}_a-\ell^{\mathrm{f}}_b|}
{\max\{1,\max_{u\in\mathcal{V}}\ell_u^{\mathrm{f}}\}},
\frac{c_v^{\mathrm{s}}-c_v^{\mathrm{o}}}
{\max\{c_v^{\mathrm{s}}+c_v^{\mathrm{o}},1\}}
\Bigg].
\end{aligned}
\label{eq:reconv-feature}
\end{equation}
where $m_v=\mathbf{1}_{\{n_v>0\}}$ denotes the first entry.
The six entries of $\mathbf{r}_v$ encode reconvergence presence,
Jaccard support overlap, the overlap coefficient, log-scaled shared-support
size, normalized fanin-depth imbalance, and the contrast between same- and
opposite-parity overlap.
Because a PI may appear in both parity-indexed support sets,
$c_v^{\mathrm{s}}$ and $c_v^{\mathrm{o}}$ are overlap statistics rather than
mutually exclusive counts.
For nodes other than two-input AND gates, we set
$\mathbf{r}_v=\mathbf{0}$ and $m_v=0$.
The descriptor therefore characterizes both the existence and the extent of
shared upstream dependency, together with its structural and polarity-related
properties.

An initialization layer constructs the local node representation from
signed-fanin neighborhoods.
We denote the resulting state at layer $l$ as
$\mathbf{h}_v^{(l)}\in\mathbb{R}^{2d_h}$.
For each relation $r\in\{0,1\}$, we construct a relation-aware message
from the corresponding fanin neighborhood:
\begin{equation}
\mathbf{m}_{v,r}^{(l)}
=
\operatorname{Agg}_{r}^{(l)}
\left(
\{\mathbf{h}_{u}^{(l)} \mid u\in\mathcal{N}_{r}(v)\}
\right),
\qquad
\mathbf{a}_{v}^{(l)}
=
\left[
\mathbf{m}_{v,0}^{(l)}
\mathbin{\Vert}
\mathbf{m}_{v,1}^{(l)}
\right].
\label{eq:relation-aggregation}
\end{equation}
where $\operatorname{Agg}_{r}^{(l)}(\cdot)$ denotes the relation-aware
aggregation over fanins connected through edge relation $r$, with empty
neighborhoods mapped to zero.

Using the PI-dependency representation $\mathbf{e}_v$ introduced in
Section~\ref{sec:support}, we jointly compute the dependency gate and the
updated node representation as
\begin{equation}
\begin{aligned}
\mathbf{g}_v^{(l)}
&=
m_v\,\sigma\!\Big(
\mathrm{MLP}_g
\big[
\mathbf{r}_v
\mathbin{\Vert}
\mathrm{MLP}_s(\mathbf{a}_v^{(l)})
\big]
\Big),\\
\mathbf{h}_v^{(l+1)}
&=
\tanh\!\Big(
\mathbf{W}_h
\big[
\mathbf{a}_v^{(l)}
\mathbin{\Vert}
\mathbf{h}_v^{(l)}
\mathbin{\Vert}
\mathbf{e}_v
\mathbin{\Vert}
\bar{\mathbf{g}}_v^{(l)}\odot\mathbf{e}_v
\big]
\Big).
\end{aligned}
\label{eq:reconv-update}
\end{equation}
where
$\mathbf{g}_v^{(l)}\in\mathbb{R}^{d_h}$ is the dependency gate and
$\bar{\mathbf{g}}_v^{(l)}
=
[\mathbf{g}_v^{(l)}\mathbin{\Vert}\mathbf{g}_v^{(l)}]
\in\mathbb{R}^{2d_h}$
broadcasts the gate to the PI-dependency representation,
$\mathrm{MLP}_s$ and $\mathrm{MLP}_g$ are multilayer perceptrons,
$\mathbf{W}_h$ denotes the state projection,
$\sigma$ is the sigmoid function, and $\odot$ denotes element-wise
multiplication.
The base PI-dependency representation $\mathbf{e}_v$ is retained for every
node, while the additional gated residual
$\bar{\mathbf{g}}_v^{(l)}\odot\mathbf{e}_v$
is activated when shared upstream support is detected.

Therefore, exact PI-support overlap serves as an explicit dependency signal
that adaptively controls how global PI information participates in local
propagation at reconvergent nodes while preserving the original AIG topology.

\subsection{Training Objective}
\label{sec:loss}
To improve representation consistency under synthesis-induced topology
changes, we augment task supervision with logic-inspired regularization.
Since Boolean operator semantics remain valid across different structural
realizations, we impose NOT complementarity and feasible probability
bounds for AND gates.
Given the final node representation $\mathbf{h}_v^{(L)}$, we predict
$p_v=\sigma(f_p(\tanh(\mathbf{W}_z\mathbf{h}_v^{(L)})))$.
The task loss $\mathcal{L}_{\mathrm{task}}$ is detailed in
Appendix~\ref{app:task-supervision}.

\paragraph{\textbf{Logic-Inspired Constraints.}}
For an AND input relation $(u,o)$, let
$\widetilde{p}_{uo}=p_u$ for a non-inverting relation and
$\widetilde{p}_{uo}=1-p_u$ for an inverting relation.
For a two-input AND node $o$ with fanins $a$ and $b$, the corresponding
Fr\'echet bounds are
\begin{equation}
\underline{p}_o
=
[\widetilde{p}_{ao}+\widetilde{p}_{bo}-1]_+,
\qquad
\overline{p}_o
=
\min(\widetilde{p}_{ao},\widetilde{p}_{bo}),
\label{eq:logic-bounds}
\end{equation}
where $[x]_+=\max(x,0)$.
These bounds require no fanin-independence assumption and therefore
remain applicable to reconvergent structures.

We penalize violations of NOT complementarity and AND probability
bounds through
\begin{equation}
\begin{aligned}
\mathcal{L}_{\mathrm{NOT}}
&=
\frac{1}{|\mathcal{E}_{\mathrm{NOT}}|}
\sum_{(u,v)\in\mathcal{E}_{\mathrm{NOT}}}
|p_u+p_v-1|,
\\
\mathcal{L}_{\mathrm{AND}}
&=
\frac{1}{|\mathcal{A}|}
\sum_{o\in\mathcal{A}}
\Big(
[\underline{p}_o-p_o]_+
+
[p_o-\overline{p}_o]_+
\Big),
\end{aligned}
\label{eq:logic-loss}
\end{equation}
where $\mathcal{E}_{\mathrm{NOT}}$ contains the input--output node
pairs of explicit NOT operations, and $\mathcal{A}$ is the set of
two-input AND nodes.

\paragraph{\textbf{Overall Objective.}}
The complete training objective is
\begin{equation}
\mathcal{L}_{\mathrm{train}}
=
\mathcal{L}_{\mathrm{task}}
+
\lambda_{\mathrm{logic}}
\Big(
\lambda_{\mathrm{NOT}}\mathcal{L}_{\mathrm{NOT}}
+
\lambda_{\mathrm{AND}}\mathcal{L}_{\mathrm{AND}}
\Big),
\label{eq:total-loss}
\end{equation}
where $\lambda_{\mathrm{logic}}$ controls the overall regularization
strength, while $\lambda_{\mathrm{NOT}}$ and $\lambda_{\mathrm{AND}}$
balance the two logic terms.
Together, these terms regularize local Boolean probability relations
that remain applicable across different structural realizations,
reducing reliance on topology-specific patterns.

\section{Experiments}
\label{sec:experiments}

\subsection{Experimental Settings}
\label{sec:experimental-settings}

\begingroup
\setlength{\emergencystretch}{1em}
\textbf{Datasets.}
We evaluate CircuitGate on three circuit benchmarks:
\textbf{ForgeEDA}~\cite{shi2025forgeeda},
\textbf{EPFL}~\cite{amaru2015epfl},
and \textbf{ITC'99}~\cite{corno2000rt}.
ForgeEDA contains 83,155 circuit instances derived from
4,450 AIGs across 1,189 open-source designs.
We use a source-disjoint split for ForgeEDA and parent-circuit-isolated
splits for EPFL and ITC'99 to prevent data leakage across splits.
Detailed dataset statistics and split settings are provided in
Appendix~\ref{app:dataset-statistics}.

\textbf{Baselines.}
We compare CircuitGate with general-purpose GNNs
(GCN~\cite{kipf2016semi}, GAT~\cite{velivckovic2017graph},
GraphSAGE~\cite{hamilton2017inductive}),
the DeepGate series~\cite{li2022deepgate,shi2023deepgate2,
shi2024deepgate3,zheng2025deepgate4},
and recent circuit-specific methods
(HOGA~\cite{deng2024less}, PolarGate~\cite{liu2024polargate},
FuncGNN~\cite{zhao2026funcgnn}, AIGer~\cite{sun2025aiger},
MGVGA~\cite{wu2025circuit}).

\textbf{Evaluation Metrics.}
For equivalent-gate identification, MAE and MSE evaluate graded functional
similarity against the continuous target $1-d_{ij}^{\mathrm{tt}}$, whereas
AP and F1 evaluate exact equivalence defined by $d_{ij}^{\mathrm{tt}}=0$,
with MAE as the primary metric.
For signal-probability prediction, we report MAE.
For topology robustness, we use CKA to measure representation consistency
between matched nodes before and after synthesis transformations.
For efficiency, we report parameter count, inference latency, and
accuracy--efficiency cost (AEC).
Detailed metric definitions, class statistics, and evaluation protocols are
provided in Appendix~\ref{app:evaluation-protocol}.

\begin{table*}[!t]
\centering
\captionsetup{
    font=small,
    justification=centering,
    singlelinecheck=false,
    skip=3pt
}
\caption{Equivalent-gate identification results.
Best and second-best results are bold and underlined.}
\label{tab:eq_results}
\scriptsize
\renewcommand{\arraystretch}{0.9}
\setlength{\tabcolsep}{2pt}
\begin{tabularx}{\textwidth}{
@{}l c C C C C C C C C C C C C@{}
}
\toprule
\multirow{2}{*}{Model}
& \multirow{2}{*}{Venue}
& \multicolumn{4}{c}{ForgeEDA}
& \multicolumn{4}{c}{EPFL}
& \multicolumn{4}{c}{ITC'99} \\
\cmidrule(lr){3-6}
\cmidrule(lr){7-10}
\cmidrule(lr){11-14}
&
& MAE $\downarrow$
& MSE $\downarrow$
& AP $\uparrow$
& F1 $\uparrow$
& MAE $\downarrow$
& MSE $\downarrow$
& AP $\uparrow$
& F1 $\uparrow$
& MAE $\downarrow$
& MSE $\downarrow$
& AP $\uparrow$
& F1 $\uparrow$ \\
\midrule
GCN
& ICLR'17
& 0.1522 & 0.0537 & 0.0141 & 0.0368
& 0.2364 & 0.1060 & 0.2952 & 0.1478
& 0.2129 & 0.1145 & 0.6396 & 0.6434 \\
GAT
& ICLR'18
& 0.1886 & 0.0762 & 0.0102 & 0.0312
& 0.3086 & 0.1517 & 0.1694 & 0.2930
& 0.2186 & 0.1205 & 0.5370 & 0.6873 \\
GraphSAGE
& NeurIPS'17
& 0.1311 & 0.0424 & 0.0175 & 0.0228
& 0.1117 & 0.0368 & 0.4118 & 0.4062
& 0.2188 & 0.1203 & 0.5501 & 0.6883 \\
\midrule
DeepGate
& DAC'22
& 0.1440 & 0.0526 & 0.0108 & 0.0296
& 0.1183 & 0.0368 & 0.4042 & 0.2705
& 0.1318 & 0.0516 & 0.7168 & 0.8133 \\
DeepGate2
& ICCAD'23
& 0.2914 & 0.1237 & 0.0154 & 0.0360
& 0.2804 & 0.1227 & 0.1316 & 0.0109
& 0.2173 & 0.0977 & 0.6717 & 0.7511 \\
DeepGate3
& ICCAD'24
& \underline{0.0553}
& \underline{0.0167}
& \underline{0.0335}
& 0.0672
& 0.1115
& 0.0361
& \underline{0.6421}
& 0.4027
& 0.0748
& 0.0206
& \underline{0.9507}
& \underline{0.9290} \\
DeepGate4
& ICLR'25
& 0.1361
& 0.0334
& 0.0305
& 0.0749
& 0.0863
& \underline{0.0217}
& 0.5621
& \underline{0.5705}
& 0.0755
& \underline{0.0197}
& 0.9058
& 0.9264 \\
\midrule
HOGA
& DAC'24
& 0.1567 & 0.0547 & 0.0149 & 0.0397
& 0.1278 & 0.0397 & 0.3551 & 0.3855
& 0.2407 & 0.1570 & 0.7926 & 0.8229 \\
PolarGate
& ICCAD'24
& 0.1567 & 0.0593 & 0.0133 & 0.0340
& 0.1450 & 0.0529 & 0.2772 & 0.1468
& 0.1135 & 0.0444 & 0.8009 & 0.8719 \\
FuncGNN
& TRETS'26
& 0.1075 & 0.0302 & 0.0180 & 0.0498
& 0.0952 & 0.0250 & 0.4918 & 0.3599
& 0.1014 & 0.0342 & 0.8736 & 0.8447 \\
AIGer
& TCAD'26
& 0.1472 & 0.0548 & 0.0180 & 0.0454
& 0.1432 & 0.0509 & 0.2761 & 0.3806
& 0.1160 & 0.0411 & 0.8701 & 0.8905 \\
MGVGA
& ICLR'25
& 0.0563 & 0.0173 & 0.0264 & \underline{0.0759}
& \underline{0.0848}
& 0.0228
& 0.5187
& 0.4170
& \underline{0.0718}
& 0.0205
& 0.9101
& 0.9210 \\
\midrule
\textbf{CircuitGate}
&
& \textbf{0.0433}
& \textbf{0.0118}
& \textbf{0.0469}
& \textbf{0.1149}
& \textbf{0.0683}
& \textbf{0.0163}
& \textbf{0.6503}
& \textbf{0.6543}
& \textbf{0.0622}
& \textbf{0.0140}
& \textbf{0.9823}
& \textbf{0.9467} \\
\bottomrule
\end{tabularx}
\vspace{-10pt}
\end{table*}

\textbf{Implementation Details.}
All experiments are conducted on an NVIDIA RTX A6000 GPU with a fixed random seed of 2024.
For fair comparison, all methods use the same dataset splits and evaluation protocols.
For each baseline, we follow its official implementation and hyperparameter settings whenever available; when adaptation is required, hyperparameters are selected using only the validation set.
CircuitGate is optimized using Adam~\cite{kingma2015adam} with a fixed learning rate of $10^{-2}$, a weight decay of $10^{-3}$, and a batch size of 256, and is trained for 300 epochs.

\subsection{RQ1: How Does CircuitGate Compare with Existing Methods?}\label{sec:rq-main}

\textbf{Motivation.}
We compare CircuitGate with GNN-based and circuit-specific
methods on equivalent-gate identification and signal-probability prediction
across ForgeEDA, EPFL, and ITC'99.

\textbf{Results.}
As shown in Table~\ref{tab:eq_results}, CircuitGate achieves the best
equivalent-gate identification performance across all three benchmarks.

\begin{wraptable}{r}{0.50\columnwidth}
\vspace{-0.4\baselineskip}
\centering
\captionsetup{
    font=small,
    justification=raggedright,
    singlelinecheck=false,
    skip=2pt
}
\caption{Signal-probability prediction results (MAE).}
\label{tab:prob_results}
\scriptsize
\renewcommand{\arraystretch}{0.90}
\setlength{\tabcolsep}{0pt}
\begin{tabular}{@{}
>{\raggedright\arraybackslash}p{0.28\linewidth}
>{\centering\arraybackslash}p{0.24\linewidth}
>{\centering\arraybackslash}p{0.24\linewidth}
>{\centering\arraybackslash}p{0.24\linewidth}
@{}}
\toprule
Model & ForgeEDA & EPFL & ITC'99 \\
\midrule
GCN       & 0.0422 & 0.1637 & 0.1688 \\
GAT       & 0.0396 & 0.1386 & 0.0879 \\
GraphSAGE & 0.0446 & \underline{0.0536} & 0.0231 \\ \midrule
DeepGate  & 0.1604 & 0.0910 & 0.0831 \\
DeepGate2 & 0.1408 & 0.0923 & 0.0717 \\
DeepGate3 & 0.0315 & 0.0625 & 0.0161 \\
DeepGate4 & \underline{0.0246} & 0.0560 & \underline{0.0154} \\ \midrule
HOGA      & 0.0991 & 0.1046 & 0.0457 \\
PolarGate & 0.1070 & 0.0987 & 0.0349 \\
FuncGNN   & 0.0558 & 0.0770 & 0.0294 \\
AIGer     & 0.0360 & 0.0812 & 0.0559 \\
MGVGA     & 0.0279 & 0.0626 & 0.0181 \\
\midrule
\textbf{CircuitGate}
& \textbf{0.0212}
& \textbf{0.0460}
& \textbf{0.0141} \\
\bottomrule
\end{tabular}
\vspace{-0.4\baselineskip}
\end{wraptable}

On ForgeEDA, CircuitGate achieves an MAE of $0.0433$, reducing error by
21.7\% over DeepGate3, with corresponding MSE and AP improvements of
29.3\% and 40.0\%, respectively.
It also improves F1 by 51.4\% over MGVGA, the second-best method on
this metric.
On EPFL and ITC'99, it achieves MAEs of $0.0683$ and $0.0622$,
outperforming the second-best MGVGA by 19.5\% and 13.4\%, respectively,
while also attaining the best MSE, AP, and F1.
As shown in Table~\ref{tab:prob_results}, CircuitGate achieves the lowest
signal-probability MAE on all three benchmarks, reducing error by 13.8\%,
14.2\%, and 8.4\%, respectively, relative to the best baseline on each
benchmark.
Under direct transfer from ForgeEDA to OpenABC-D (OpenABC)~\cite{chowdhury2021openabc}
without fine-tuning, it also achieves the lowest equivalent-gate MAE;
detailed results are provided in
Appendix~\ref{app:cross-dataset-generalization}.

\begin{table*}[!h]
\centering
\captionsetup{
    font=small,
    justification=centering,
    singlelinecheck=false,
    skip=3pt
}
\caption{Ablation study on the equivalent-gate identification task on ForgeEDA, EPFL, and ITC'99.}
\label{tab:ablation}

\scriptsize
\renewcommand{\arraystretch}{0.9}
\setlength{\tabcolsep}{2pt}

\begin{tabularx}{\textwidth}{
@{}l C C C C C C C C C C C C@{}
}
\toprule
\multirow{2}{*}{Variant}
& \multicolumn{4}{c}{ForgeEDA}
& \multicolumn{4}{c}{EPFL}
& \multicolumn{4}{c}{ITC'99} \\
\cmidrule(lr){2-5}
\cmidrule(lr){6-9}
\cmidrule(lr){10-13}
& MAE $\downarrow$
& MSE $\downarrow$
& AP $\uparrow$
& F1 $\uparrow$
& MAE $\downarrow$
& MSE $\downarrow$
& AP $\uparrow$
& F1 $\uparrow$
& MAE $\downarrow$
& MSE $\downarrow$
& AP $\uparrow$
& F1 $\uparrow$ \\
\midrule
w/o Circuit-Level Context
& 0.0477
& 0.0135
& 0.0429
& 0.1118
& 0.0914
& 0.0234
& 0.4409
& 0.4403
& 0.0850
& 0.0243
& 0.9755
& 0.9270 \\
w/o PI-Dependency Encoding
& 0.0582
& 0.0184
& 0.0220
& 0.0634
& 0.0864
& 0.0248
& 0.5211
& 0.3708
& 0.0691
& 0.0210
& 0.9454
& 0.9317 \\
w/o Reconvergence 
& 0.0561
& 0.0175
& 0.0229
& 0.0670
& 0.0872
& 0.0243
& 0.5472
& 0.3964
& 0.0833
& 0.0256
& 0.9661
& 0.9084 \\
w/o Logic Constraints
& 0.0495
& 0.0135
& 0.0351
& 0.1007
& 0.0817
& 0.0220
& 0.5203
& 0.5513
& 0.0864
& 0.0201
& 0.9659
& 0.8985 \\
\midrule
\textbf{Full}
& \textbf{0.0433}
& \textbf{0.0118}
& \textbf{0.0469}
& \textbf{0.1149}
& \textbf{0.0683}
& \textbf{0.0163}
& \textbf{0.6503}
& \textbf{0.6543}
& \textbf{0.0622}
& \textbf{0.0140}
& \textbf{0.9823}
& \textbf{0.9467} \\
\bottomrule
\end{tabularx}
\vspace{-10pt}
\end{table*}

\subsection{RQ2: How Does Each Component Contribute to CircuitGate?}
\label{sec:rq-ablation}

\textbf{Motivation.}
We assess the contribution of each CircuitGate component.

\textbf{Results.}
As shown in Table~\ref{tab:ablation}, all components contribute to CircuitGate's
performance. Removing PI-support encoding causes the largest MAE degradation
on ForgeEDA (34.4\%), followed by reconvergence modeling (29.6\%), highlighting
the importance of global PI dependencies and reconvergent-fanin relationships.
The impact of different components varies across benchmarks: circuit-level
context is most important on EPFL, while logic constraints provide the largest
gain on ITC'99. These results demonstrate that the proposed components
complement each other in learning circuit-level functional representations.

We further evaluate the PI-support design on ForgeEDA. Per-circuit PI-token
permutation maintains comparable performance, whereas removing polarity
information or retaining only cardinality/coverage statistics leads to clear
degradation, confirming the importance of fine-grained polarity-aware PI
support. Detailed results are provided in Appendix~\ref{app:support-controls}.




\subsection{RQ3: How Efficient and Scalable Is CircuitGate?}
\label{sec:rq-efficiency}

\textbf{Motivation.}
We evaluate CircuitGate's accuracy--efficiency trade-off and preprocessing scalability.

\textbf{Results.}
As shown in Table~\ref{tab:efficiency}, on a fixed subset of 200 ForgeEDA
test circuits, CircuitGate achieves the lowest MAE of $0.0433$ with a neural
inference latency of $12.4393$\,ms, reducing MAE by 23.1\% over MGVGA with
only a 1.3\% latency increase. It also achieves the best AEC of $1.0275$.
\begin{wraptable}{r}{0.58\columnwidth}
\vspace{-0.15\baselineskip}
\centering

\captionsetup{
    font=small,
    justification=raggedright,
    singlelinecheck=false,
    skip=3pt
}
\caption{Accuracy--efficiency comparison on ForgeEDA}
\label{tab:efficiency}

\scriptsize
\renewcommand{\arraystretch}{0.90}
\setlength{\tabcolsep}{0pt}

\begin{tabular}{@{}
>{\raggedright\arraybackslash}p{0.24\linewidth}
>{\centering\arraybackslash}p{0.19\linewidth}
>{\centering\arraybackslash}p{0.19\linewidth}
>{\centering\arraybackslash}p{0.22\linewidth}
>{\centering\arraybackslash}p{0.16\linewidth}
@{}}
\toprule
Model
& \mbox{Params. (M) $\downarrow$}
& \mbox{EQ MAE $\downarrow$}
& \mbox{Neural inf. (ms) $\downarrow$}
& \mbox{AEC $\downarrow$} \\
\midrule

GraphSAGE
& \textbf{0.0173}
& 0.1311
& \textbf{9.1204}
& \underline{1.1957} \\

AIGer
& 0.5603
& 0.1472
& \underline{10.5431}
& 2.5137 \\

FuncGNN
& 0.9765
& 0.1075
& 11.2865
& 2.0854 \\

MGVGA
& 1.1050
& 0.0563
& 12.2759
& 1.2031 \\

DeepGate3
& 4.7272
& \underline{0.0553}
& 73.2356
& 8.0999 \\

DeepGate4
& 4.6465
& 0.1361
& 172.2620
& 46.8177 \\

HOGA
& \underline{0.0459}
& 0.1567
& 107.8129
& 19.8255 \\

\midrule

\textbf{CircuitGate}
& 2.8170
& \textbf{0.0433}
& 12.4393
& \textbf{1.0275} \\

\bottomrule
\end{tabular}

\vspace{-0.4\baselineskip}
\end{wraptable}
We further evaluate one-time CPU preprocessing on complete circuits across
seven graph-size ranges. On the largest graph with 525,762 nodes and
733,211 edges, feature construction takes 14.03\,s, while the complete
input-to-cache pipeline takes 14.89\,s with a peak RSS of 2.73\,GiB.
These results demonstrate that CircuitGate maintains practical preprocessing
costs on large-scale complete circuits. Detailed scaling results are provided
in Appendix~\ref{app:preprocessing-scalability}.
\subsection{RQ4: How Robust Is CircuitGate to Topology Changes?}
\label{sec:rq-robustness}

\begin{wraptable}{r}{0.58\columnwidth}
\vspace{-0.25\baselineskip}
\centering

\captionsetup{
    font=small,
    justification=raggedright,
    singlelinecheck=false,
    skip=2pt
}
\caption{Representation consistency under synthesis-induced topology transformations, measured by CKA ($\uparrow$).}
\label{tab:topology_cka_main}

\scriptsize
\renewcommand{\arraystretch}{0.82}
\setlength{\tabcolsep}{0pt}

\begin{tabular}{@{}
>{\raggedright\arraybackslash}p{0.24\linewidth}
>{\centering\arraybackslash}p{0.19\linewidth}
>{\centering\arraybackslash}p{0.19\linewidth}
>{\centering\arraybackslash}p{0.19\linewidth}
>{\centering\arraybackslash}p{0.19\linewidth}
@{}}
\toprule
Model & resyn2 & dc2 & rewrite & refactor \\
\midrule
GCN       & 0.6506 & 0.5967 & 0.7129 & 0.7322 \\
GAT       & 0.6241 & 0.5836 & 0.8174 & 0.8449 \\
GraphSAGE & 0.6631 & 0.6272 & 0.7616 & 0.7948 \\
\midrule
DeepGate  & 0.4278 & 0.4114 & 0.5219 & 0.5374 \\
DeepGate2 & 0.4637 & 0.4105 & 0.5232 & 0.5392 \\
DeepGate3 & \underline{0.6784}
          & \underline{0.6572}
          & 0.7672
          & 0.7842 \\
DeepGate4 & 0.6472 & 0.6375 & 0.7306 & 0.7498 \\
\midrule
HOGA      & 0.5840 & 0.5679 & 0.6521 & 0.6691 \\
PolarGate & 0.5934 & 0.5500 & 0.6848 & 0.7182 \\
FuncGNN   & 0.6662 & 0.6149
          & \underline{0.8471}
          & \underline{0.8490} \\
AIGer     & 0.6129 & 0.5760 & 0.7003 & 0.7319 \\
MGVGA     & 0.6099 & 0.5988 & 0.6664 & 0.6773 \\
\midrule
\textbf{CircuitGate}
& \textbf{0.7933}
& \textbf{0.7866}
& \textbf{0.8775}
& \textbf{0.8789} \\
\bottomrule
\end{tabular}

\vspace{-0.25\baselineskip}
\end{wraptable}
\textbf{Motivation.}
We examine whether CircuitGate maintains consistent representations under
functionality-preserving synthesis transformations. We evaluate all methods
on the same 20 transformed ForgeEDA test circuits using fixed checkpoints
without retraining.

\textbf{Results.}
As shown in Table~\ref{tab:topology_cka_main}, CircuitGate achieves the
highest CKA under all four synthesis transformations, with an average CKA of
$0.8341$. These results indicate that CircuitGate learns more consistent
representations under functionality-preserving topology variations.
Additional MAE results are provided in
Appendix~\ref{app:topology_robustness}.
\begin{figure}[t]
    \centering
    \includegraphics[
        width=\linewidth
    ]{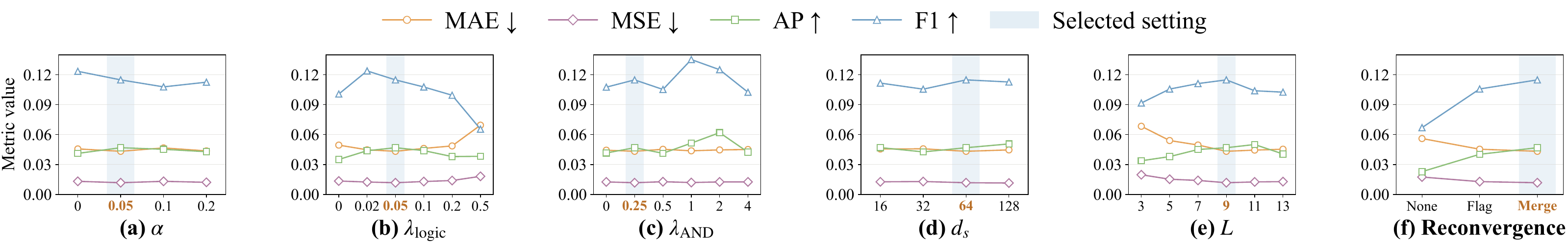}

    \captionsetup{
        font=small,
        skip=3pt
    }

\caption{
Parameter sensitivity of CircuitGate on equivalent-gate identification.
Each panel varies one hyperparameter, with the selected setting highlighted in bold orange.
}
    \label{fig:parameter-sensitivity}
    \vspace{-20pt}
\end{figure}
\subsection{RQ5: How Sensitive Is CircuitGate to Hyperparameters?}
\label{sec:rq-parameters}

\textbf{Results.}
As shown in Figure~\ref{fig:parameter-sensitivity}, the selected configuration
uses $\alpha=0.05$, $\lambda_{\mathrm{logic}}=0.05$,
$\lambda_{\mathrm{AND}}=0.25$, $\lambda_{\mathrm{NOT}}=1.0$,
$d_s=64$, $L=9$, and merge-gate reconvergence.
These settings provide a favorable operating point across the evaluated
metrics, while overly strong logic regularization or insufficient propagation
depth leads to noticeable degradation.
Further increasing the support-sketch dimension or propagation depth provides
limited additional benefit.
Detailed quantitative results are provided in
Appendix~\ref{app:RQ5}.

\section{Conclusion}

We presented CircuitGate, a framework for learning functional
representations of AIGs through circuit-level context, global
PI dependencies, and reconvergent-fanin modeling, guided by a
logic-inspired objective. These mechanisms capture functional
relationships beyond local connectivity while reducing reliance
on topology-specific structural patterns. Across the reported
ForgeEDA, EPFL, and ITC'99 evaluations, CircuitGate achieves the
lowest MAE on both equivalent-gate identification and signal-probability
prediction while maintaining competitive inference latency.
The ablation results further highlight the contributions of PI-support
encoding and reconvergence modeling, supporting the value of
circuit-level functional priors for accurate and efficient
AIG representation learning.
\subsection*{Reproducibility Statement}
We provide an anonymous implementation of CircuitGate to facilitate
reproducibility. Detailed dataset statistics and splits, evaluation protocols,
implementation settings, robustness evaluation, and hyperparameter analyses
are provided in the main text and Appendix.
\subsection*{AI Use Statement}
Generative AI tools were used solely for language polishing to improve the clarity, grammar, and readability of the manuscript. They were not used for study design, methodology development, data analysis, result generation, or scientific interpretation. All AI-assisted language edits were reviewed and approved by the authors, who take full responsibility for the final content of the manuscript.
\bibliographystyle{iclr2027_conference}
\bibliography{iclr2027_conference}

\clearpage
\appendix
\section{Additional Experimental Results}
\label{app:additional_results}
\subsection{Task Supervision Details}
\label{app:task-supervision}

For the two downstream tasks, we use the final node representation
$\mathbf{h}_v^{(L)}$ to obtain
$\mathbf{z}_v=\tanh(\mathbf{W}_z\mathbf{h}_v^{(L)})$
and predict
$p_v=\sigma(f_p(\mathbf{z}_v))$.

The two supervised losses are
\begin{equation}
\mathcal{L}_{\mathrm{prob}}
=
\frac{1}{|\mathcal{V}|}
\sum_{v\in\mathcal{V}}
|p_v-y_v|,
\qquad
\mathcal{L}_{\mathrm{eq}}
=
\frac{1}{|\mathcal{Q}|}
\sum_{(i,j)\in\mathcal{Q}}
|s_{ij}-y_{ij}^{\mathrm{eq}}|.
\label{eq:task-loss}
\end{equation}
where $\mathcal{Q}$ is the set of supervised node pairs,
$y_v$ is the ground-truth signal probability, and
$s_{ij}=\cos(\mathbf{z}_i,\mathbf{z}_j)$.

The target
$y_{ij}^{\mathrm{eq}}=1-d_{ij}^{\mathrm{tt}}$
is derived from the normalized truth-table distance
$d_{ij}^{\mathrm{tt}}$ supplied by the benchmark.
For each circuit instance, we use the 2,048 intra-instance node pairs
specified by ForgeEDA's \texttt{tt\_pair\_index};
$d_{ij}^{\mathrm{tt}}=0$ indicates functional equivalence.

For probability prediction,
$\mathcal{L}_{\mathrm{task}}=\mathcal{L}_{\mathrm{prob}}$.
For equivalent-gate identification,
\begin{equation}
\mathcal{L}_{\mathrm{task}}
=
\mathcal{L}_{\mathrm{eq}}
+
\alpha\mathcal{L}_{\mathrm{prob}},
\end{equation}
where $\alpha$ weights auxiliary probability supervision and empty-set
averages are defined as zero.
\subsection{Dataset Statistics}
\label{app:dataset-statistics}

\begin{table}[!h]
    \centering
    \captionsetup{
        font=small,
        justification=centering,
        singlelinecheck=false,
        skip=3pt
    }
    \caption{Statistics of the ForgeEDA dataset~\cite{shi2025forgeeda}.}
    \label{tab:dataset-statistics}

    \scriptsize
    \renewcommand{\arraystretch}{1.05}
    \setlength{\tabcolsep}{4pt}

    \resizebox{\linewidth}{!}{%
        \begin{tabular}{@{}lccccl@{}}
            \toprule
            Modality
            & \#Samples
            & Node Range
            & Edge Range
            & Format
            & Contents \\
            \midrule

            Code Repository
            & 1,189
            & --
            & --
            & \texttt{.v}
            & Verilog code with comments and specifications \\

            \midrule

            PM Netlist
            & 4,450
            & --
            & --
            & \texttt{.v}
            & Post-mapping netlists \\

            &
            &
            &
            &
            \texttt{.rpt}
            & Area, path-delay, and power reports \\

            \midrule

            Placed Netlist
            & 4,450
            & --
            & --
            & \texttt{.v}/\texttt{.def}
            & Placed netlists \\

            &
            &
            &
            &
            \texttt{.rpt}
            & Area, WNS, TNS, and power reports \\

            \midrule

            AIG
            & 4,450
            & [5--386,996]
            & [1--753,750]
            & \texttt{.aig}
            & Raw AIG files \\

            \midrule

            Circuit Instances
            & 83,155
            & [500--5,000]
            & [258--8,070]
            & \texttt{.aig}
            & Parsed circuit graphs \\

            &
            &
            &
            &
            \texttt{.npz}
            & PyTorch Geometric graph data \\

            \bottomrule
        \end{tabular}%
    }
\end{table}

\vspace{0.5\baselineskip}

\begin{table}[!h]
    \centering
    \captionsetup{
        font=small,
        justification=centering,
        singlelinecheck=false,
        skip=3pt
    }
    \caption{Statistics of the processed EPFL benchmark dataset.}
    \label{tab:epfl-statistics}

    \scriptsize
    \renewcommand{\arraystretch}{1.05}
    \setlength{\tabcolsep}{4pt}

    \resizebox{\linewidth}{!}{%
        \begin{tabular}{@{}lcccc l@{}}
            \toprule
            Data Type
            & \#Samples
            & Node Range
            & Edge Range
            & \#Gate Pairs
            & Contents \\
            \midrule

            Original AIGs
            & 19
            & [320--101,954]
            & [496--159,073]
            & --
            & Original EPFL circuits \\

            \midrule

            Processed Parents
            & 16
            & --
            & --
            & --
            & Parent circuits producing subcircuits \\

            \midrule

            Processed Subcircuits
            & 101
            & [519--5,000]
            & [716--9,305]
            & 1,514,460
            & Model input graphs \\

            \midrule

            Train Split
            & 51
            & [519--5,000]
            & --
            & 765,000
            & 9 parent circuits \\

            Validation Split
            & 27
            & [2,426--5,000]
            & --
            & 405,000
            & 3 parent circuits \\

            Test Split
            & 23
            & [1,082--5,000]
            & --
            & 344,460
            & 4 parent circuits \\

            \bottomrule
        \end{tabular}%
    }
\end{table}

\begin{table}[t]
    \centering
    \captionsetup{
        font=small,
        justification=centering,
        singlelinecheck=false,
        skip=3pt
    }
    \caption{Statistics of the processed ITC'99 benchmark dataset.}
    \label{tab:itc99-statistics}

    \scriptsize
    \renewcommand{\arraystretch}{1.05}
    \setlength{\tabcolsep}{3pt}

    \begin{tabularx}{\linewidth}{
        @{}
        l
        c
        c
        c
        c
        >{\raggedright\arraybackslash}X
        @{}
    }
        \toprule
        Data Type
        & \#Samples
        & Node Range
        & Edge Range
        & \#Gate Pairs
        & Contents \\
        \midrule

        Original AIGs
        & 20
        & [47--140,638]
        & [65--217,943]
        & --
        & Original ITC'99 circuits \\

        \midrule

        Processed Parents
        & 16
        & --
        & --
        & --
        & Parent circuits producing subcircuits \\

        \midrule

        Processed Subcircuits
        & 57
        & [79--5,000]
        & [113--7,321]
        & 74,906
        & Model input graphs \\

        \midrule

        Train Split
        & 38
        & [79--5,000]
        & [113--7,200]
        & 41,564
        & 10 parent circuits \\

        Validation Split
        & 7
        & [538--5,000]
        & [720--6,814]
        & 14,934
        & 3 parent circuits \\

        Test Split
        & 12
        & [1,154--5,000]
        & [1,229--7,321]
        & 18,408
        & 3 parent circuits \\

        \bottomrule
    \end{tabularx}
\end{table}

\subsection{Evaluation Protocol and Efficiency Metric}
\label{app:evaluation-protocol}

\paragraph{Dataset Splits.}
For ForgeEDA, we use a source-disjoint split containing 4,158 training,
4,158 validation, and 74,839 test circuit instances, corresponding to
5\%, 5\%, and 90\% of the dataset, respectively.
Circuit instances derived from the same source design are assigned to only
one split.

For EPFL, among the 19 original AIGs, 16 produce valid subcircuits after
preprocessing. These 16 parent circuits are partitioned into 9 training,
3 validation, and 4 test circuits, resulting in 51 training, 27 validation,
and 23 test subcircuits, respectively. Subcircuits derived from the same
parent circuit are assigned to only one split, ensuring no parent-circuit
leakage. Thus, the EPFL evaluation measures performance on held-out EPFL
circuits rather than zero-shot transfer from ForgeEDA.

For ITC'99, the processed dataset contains 57 subcircuits derived from
16 parent circuits. We use 38 training, 7 validation, and 12 test
subcircuits from 10, 3, and 3 mutually disjoint parent circuits,
respectively. No original parent circuit contributes subcircuits to more
than one split.

For all three benchmarks, checkpoints are selected using validation MAE,
and the test splits are used only for final evaluation.
For ITC'99, equivalent-gate identification and signal-probability prediction
are trained independently on the corresponding training split, with
checkpoint selection performed on the validation split before test evaluation.

\paragraph{Equivalent-Gate Metrics.}
For a gate pair $(i,j)$, the predicted score is
$s_{ij}=\cos(\mathbf{z}_i,\mathbf{z}_j)$, and the continuous target is
$t_{ij}=1-d_{ij}^{\mathrm{tt}}$.
A pair is treated as positive when $d_{ij}^{\mathrm{tt}}=0$ and negative
otherwise.
Mean absolute error (MAE) and mean squared error (MSE) measure the difference
between $s_{ij}$ and $t_{ij}$.
Average precision (AP) is computed directly from the continuous prediction
scores.
For F1, the classification threshold is selected on the validation split by
maximizing validation F1 and is then fixed for test evaluation.
MAE and MSE are first computed within each circuit and then averaged across
test circuits, whereas AP and F1 are computed from the pooled test pairs.
\paragraph{Representation Consistency Metric.}
For topology robustness, we use linear centered kernel alignment (CKA) to
measure the consistency of node representations before and after synthesis
transformations. Given centered representation matrices
$\mathbf{X}$ and $\mathbf{Y}$, linear CKA is defined as
\begin{equation}
\mathrm{CKA}(\mathbf{X},\mathbf{Y})
=
\frac{
\|\mathbf{X}^{\top}\mathbf{Y}\|_{F}^{2}
}{
\|\mathbf{X}^{\top}\mathbf{X}\|_{F}
\,
\|\mathbf{Y}^{\top}\mathbf{Y}\|_{F}
},
\label{eq:cka}
\end{equation}
where $\|\cdot\|_{F}$ denotes the Frobenius norm.
A higher CKA indicates greater consistency between the learned
representations under synthesis-induced topology transformations.
\paragraph{Pair Distribution.}
Table~\ref{tab:eq-pair-distribution} reports the prevalence of
exact-equivalence pairs in each test set. ForgeEDA is highly imbalanced,
with only $0.464648\%$ of evaluated pairs satisfying
$d_{ij}^{\mathrm{tt}}=0$. The corresponding positive-pair rate is
$10.4561\%$ on EPFL, while ITC'99 uses balanced positive and negative
pairs. Since AP depends on class prevalence, its absolute values should
be interpreted together with the positive-pair rate of each benchmark.

\begin{table}[t]
\centering
\caption{Exact-equivalence pair statistics on the test sets.}
\label{tab:eq-pair-distribution}
\scriptsize

\begin{tabular*}{\linewidth}{@{\extracolsep{\fill}}lrrr@{}}
\toprule
Dataset
& Evaluated Pairs
& Positive Pairs
& Positive Rate \\
\midrule

ForgeEDA
& 153,270,272
& 712,168
& 0.464648\% \\

EPFL
& 344,460
& 36,017
& 10.4561\% \\

ITC'99
& 18,408
& 9,204
& 50.0000\% \\

\bottomrule
\end{tabular*}
\end{table}

\paragraph{Latency Measurement.}
Inference latency is measured on a fixed subset of 200 test circuits using
an NVIDIA RTX A6000 GPU.
Each model is evaluated on the same circuits with a batch size of one.
The reported latency measures neural inference on preprocessed circuit graphs
and excludes one-time graph preprocessing and PI-support construction.

\paragraph{Accuracy--Efficiency Cost.}
To jointly evaluate predictive accuracy, inference latency, and model
complexity, we define the accuracy--efficiency cost (AEC) for model $i$ as
\begin{equation}
\mathrm{AEC}_i
=
\mathrm{MAE}_i \cdot \mathrm{Latency}_i
\left(
1+
\frac{\log(N_i/N_{\min})}
{\log(N_{\max}/N_{\min})}
\right),
\end{equation}
where $\mathrm{Latency}_i$ is measured in milliseconds,
$N_i$ is the number of trainable parameters of model $i$, and
$N_{\min}$ and $N_{\max}$ are the minimum and maximum parameter counts
among the models compared in Table~\ref{tab:efficiency}.
Lower AEC indicates a better joint trade-off among prediction accuracy,
inference latency, and model complexity.
The logarithmic normalization maps the model-size penalty to $[0,1]$,
so the complexity term scales
$\mathrm{MAE}\times\mathrm{Latency}$ by a factor between one and two.
\begin{table}[h]
\centering
\caption{Additional controls for PI-support encoding on ForgeEDA.}
\label{tab:support-controls}
\scriptsize
\renewcommand{\arraystretch}{0.95}

\begin{tabular*}{\linewidth}{@{\extracolsep{\fill}}lcccc@{}}
\toprule
Variant
& MAE $\downarrow$
& MSE $\downarrow$
& AP $\uparrow$
& F1 $\uparrow$ \\
\midrule

Full
& 0.04330
& 0.01180
& 0.04690
& 0.11490 \\

Per-circuit PI permutation
& 0.04281
& 0.01204
& 0.04540
& 0.11447 \\

Unsigned support
& 0.04416
& 0.01287
& 0.04205
& 0.05849 \\

Cardinality/Coverage only
& 0.05360
& 0.01724
& 0.02781
& 0.07850 \\

\bottomrule
\end{tabular*}
\end{table}

\subsection{Cross-Dataset Generalization to OpenABC}
\label{app:cross-dataset-generalization}

To further evaluate cross-dataset generalization, we directly apply the
ForgeEDA-trained checkpoints to OpenABC without any fine-tuning or parameter
adaptation. All methods use their corresponding checkpoints trained on
ForgeEDA and are evaluated under the same OpenABC test protocol. This setting
therefore measures how well the learned circuit representations transfer to
an unseen circuit distribution.

As shown in Figure~\ref{fig:openabc-generalization}, CircuitGate achieves the
lowest MAE of $0.10125$ under direct ForgeEDA-to-OpenABC transfer, followed by
DeepGate3 with an MAE of $0.10223$. CircuitGate also outperforms MGVGA
($0.11177$), DeepGate4 ($0.11476$), and the remaining baselines under the same
zero-shot evaluation setting. These results provide additional evidence that
the functional representations learned by CircuitGate transfer effectively
across circuit datasets without target-domain adaptation.

\subsection{Additional PI-Support Controls}
\label{app:support-controls}
\begin{wrapfigure}{r}{0.48\linewidth}
    \centering
    \includegraphics[
        width=\linewidth
    ]{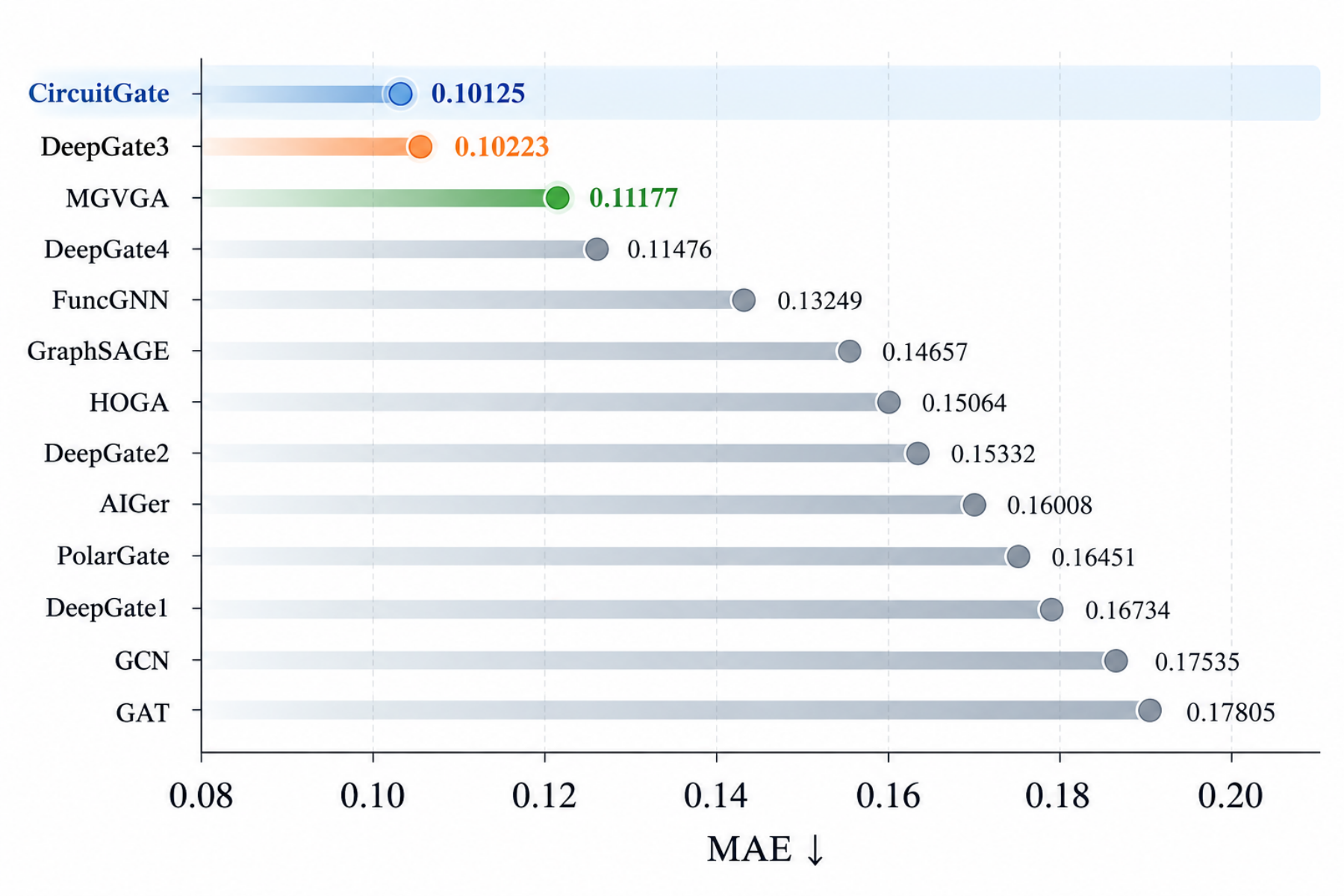}

    \captionsetup{
        font=small,
        justification=raggedright,
        singlelinecheck=false,
        skip=3pt
    }

    \caption{
    Cross-dataset generalization from ForgeEDA to OpenABC.
    All methods are directly evaluated on OpenABC using checkpoints trained
    on ForgeEDA, without fine-tuning. Lower MAE is better.
    }
    \label{fig:openabc-generalization}
\end{wrapfigure}
To further examine the PI-support encoding, we compare the full model with
three controlled variants on ForgeEDA. Per-circuit PI-token permutation
independently changes the correspondence between graph-local PI indices and
random tokens within each circuit. Unsigned support removes parity separation,
while cardinality/coverage-only encoding retains only coarse support statistics.

The per-circuit permutation control closely matches the full model, providing
no evidence that CircuitGate relies on cross-circuit PI-index alignment.
In contrast, removing support polarity or reducing the representation to
cardinality and coverage statistics degrades performance, indicating that the
main benefit comes from fine-grained, polarity-aware support structure.

\subsection{Complete-Graph Preprocessing Scalability}
\label{app:preprocessing-scalability}

We further evaluate the scalability of CircuitGate's one-time preprocessing
pipeline on complete OpenABC circuits. The circuit inventory is partitioned
into seven intervals according to the resulting CircuitGate graph size:
$[500,2\mathrm{k})$, $[2\mathrm{k},8\mathrm{k})$,
$[8\mathrm{k},32\mathrm{k})$, $[32\mathrm{k},128\mathrm{k})$,
$[128\mathrm{k},256\mathrm{k})$, $[256\mathrm{k},512\mathrm{k})$,
and $[512\mathrm{k},+\infty)$. The largest complete graph in each interval
is selected before profiling. No graph is truncated, and runtime or memory
measurements are not used during circuit selection.

Each circuit is evaluated in three independent fresh processes with newly
generated output caches, and Table~\ref{tab:preprocessing-scalability}
reports the median results. Preprocessing includes parity-specific PI-support
construction, support-sketch generation, reconvergence descriptors,
EDA-native feature construction, and cache serialization. The
\emph{input-to-cache} time additionally includes raw input parsing, but does
not include neural inference. Peak RSS denotes the peak resident memory of
the complete preprocessing process, including the Python, NumPy, PyTorch,
and framework base memory footprint. All measurements are conducted on an
Intel Core i9-13900K CPU with 128\,GiB RAM.

\begin{table*}[h]
\centering
\caption{
Complete-graph preprocessing scalability. $M$ denotes the total
number of parity-specific support memberships. Values are medians of three
independent fresh-process runs.
}
\label{tab:preprocessing-scalability}

\scriptsize
\renewcommand{\arraystretch}{0.95}

\begin{tabular*}{\textwidth}{@{\extracolsep{\fill}}lrrrrrrr@{}}
\toprule
Circuit
& Nodes
& Edges
& PIs
& $M$
& Preproc. (s)
& Input-to-cache (s)
& Peak RSS (MiB) \\
\midrule

\texttt{simple\_spi}
& 1,928
& 2,694
& 164
& 15,984
& 0.0350
& 0.0535
& 362.49 \\

\texttt{max}
& 6,279
& 8,632
& 512
& 2,373,500
& 0.3432
& 0.3691
& 368.83 \\

\texttt{mem\_ctrl}
& 31,001
& 47,906
& 1,187
& 2,465,821
& 0.9151
& 0.9828
& 419.16 \\

\texttt{tinyRocket}
& 104,336
& 152,090
& 4,561
& 8,373,687
& 3.3765
& 3.5616
& 566.09 \\

\texttt{jpeg}
& 233,573
& 343,382
& 4,962
& 4,816,133
& 4.6663
& 5.0562
& 857.41 \\

\texttt{hyp}
& 420,769
& 634,848
& 256
& 88,956,323
& 15.0327
& 15.7461
& 885.24 \\

\texttt{dft}
& 525,762
& 733,211
& 37,597
& 5,381,365
& 14.0344
& 14.8853
& 2,796.00 \\

\bottomrule
\end{tabular*}
\end{table*}

All 21 profiling runs complete successfully. The largest evaluated graph,
\texttt{dft}, contains 525,762 nodes and 733,211 edges, for which mandatory
feature construction takes 14.03\,s and the full input-to-cache pipeline
takes 14.89\,s with a peak RSS of 2.73\,GiB. Runtime depends not only on
graph size but also on support density; for example, \texttt{hyp} contains
fewer nodes than \texttt{dft} but requires 15.03\,s of preprocessing due to
its 88.96 million parity-specific support memberships. These results
demonstrate that complete-graph preprocessing remains feasible at the tested
scale.

\subsection{Synthesis-Induced Topology Robustness Details}
\label{app:topology_robustness}

For the topology-robustness evaluation, we randomly select a fixed subset of
20 eligible ForgeEDA test circuits and independently generate
\texttt{resyn2}, \texttt{dc2}, \texttt{rewrite}, and \texttt{refactor}
views for each circuit. Whole-circuit functional equivalence between the
original and transformed AIGs is first verified using ABC CEC.

To establish cross-view node correspondences, each graph is simulated using
15,000 identical random input patterns under the same PI ordering and a fixed
seed. Nodes with identical simulation signatures are treated as candidate
matches, and the resulting mappings are further verified by CEC.
All methods are evaluated using fixed ForgeEDA checkpoints without retraining.

For representation consistency, we compute CKA between the representations
of matched nodes in the original and transformed graphs. The corresponding
results are reported in Table~\ref{tab:topology_cka_main}.

\paragraph{Task-Level Robustness.}
We additionally evaluate equivalent-gate MAE on the same transformed circuits.
Original-graph EQ pairs are retained only when both endpoints can be mapped
in all four transformed views. The transformed views reuse the corresponding
original truth-table distances $d_{ij}^{\mathrm{tt}}$, and all methods are
evaluated on the same matched pairs.

\begin{table}[t]
\centering
\captionsetup{
    font=small,
    justification=centering,
    singlelinecheck=false,
    skip=3pt
}
\caption{
Equivalent-gate identification MAE ($\downarrow$) under
synthesis-induced topology transformations.
Best and second-best results are bold and underlined, respectively.
}
\label{tab:topology_mae_appendix}

\small
\renewcommand{\arraystretch}{0.95}
\setlength{\tabcolsep}{8pt}

\begin{tabular*}{\linewidth}{@{\extracolsep{\fill}}lcccc@{}}
\toprule
Model & resyn2 & dc2 & rewrite & refactor \\
\midrule
GCN       & 0.3724 & 0.3721 & 0.3732 & 0.3726 \\
GAT       & 0.3662 & 0.3696 & 0.3649 & 0.3644 \\
GraphSAGE & 0.2477 & 0.2417 & 0.2447 & 0.2444 \\
\midrule
DeepGate  & 0.2138 & 0.2275 & 0.2474 & 0.2395 \\
DeepGate2 & 0.2814 & 0.2805 & 0.2850 & 0.2911 \\
DeepGate3 & \underline{0.1971}
          & \underline{0.1933}
          & \underline{0.1919}
          & \underline{0.1921} \\
DeepGate4 & 0.2361 & 0.2335 & 0.2366 & 0.2380 \\
\midrule
HOGA      & 0.3806 & 0.3771 & 0.3759 & 0.3754 \\
PolarGate & 0.2580 & 0.2603 & 0.2627 & 0.2625 \\
FuncGNN   & 0.2537 & 0.2573 & 0.2583 & 0.2571 \\
AIGer     & 0.2499 & 0.2523 & 0.2567 & 0.2565 \\
MGVGA     & 0.2249 & 0.2165 & 0.2180 & 0.2186 \\
\midrule
\textbf{CircuitGate}
& \textbf{0.1661}
& \textbf{0.1630}
& \textbf{0.1138}
& \textbf{0.1152} \\
\bottomrule
\end{tabular*}
\end{table}

CircuitGate achieves the lowest MAE under all four transformations,
providing complementary task-level evidence for its robustness to
synthesis-induced topology variations.

\subsection{Additional Materials of RQ5: Hyperparameter Sensitivity}
\label{app:RQ5}

\textbf{Motivation.}
We further examine the sensitivity of CircuitGate to its loss weights,
support-sketch dimension, propagation depth, and reconvergence formulation.
For each parameter group, we vary one setting while keeping the remaining
components fixed.
The default configuration is
$\alpha=0.05$,
$\lambda_{\mathrm{logic}}=0.05$,
$\lambda_{\mathrm{AND}}=0.25$,
$\lambda_{\mathrm{NOT}}=1.0$,
$d_s=64$,
$L=9$,
with merge-gate reconvergence.

\textbf{Results.}
Table~\ref{tab:parameter-sensitivity} reports the detailed results.
With MAE as the primary metric, the default settings
$\alpha=0.05$,
$\lambda_{\mathrm{logic}}=0.05$,
$\lambda_{\mathrm{AND}}=0.25$,
$d_s=64$, and $L=9$
achieve the lowest MAE within their corresponding parameter groups.

For $\alpha$, MAE varies from $0.0433$ to $0.0467$ across the evaluated
settings. For $\lambda_{\mathrm{logic}}$, the default value of $0.05$
achieves an MAE of $0.0433$, whereas increasing the weight to $0.5$
degrades MAE to $0.0695$. Similarly, $\lambda_{\mathrm{AND}}=0.25$
achieves an MAE of $0.0433$, while increasing it to $4.0$ raises MAE
to $0.0451$.

For the architectural parameters, $d_s=64$ provides the lowest MAE among
the evaluated support dimensions. Increasing the propagation depth from
three to nine layers progressively improves MAE from $0.0684$ to $0.0433$,
whereas further increasing the depth to 11 or 13 layers does not provide
additional gains. For reconvergence modeling, the merge-gate formulation
achieves an MAE of $0.0433$, reducing MAE by 4.6\% relative to the binary
flag formulation and by 22.8\% relative to omitting reconvergence modeling.

Overall, the results show that CircuitGate performs favorably across moderate
parameter settings, while overly strong logic regularization or insufficient
propagation depth leads to noticeable performance degradation.
\begin{table}[t]
\centering
\captionsetup{
    font=small,
    justification=centering,
    singlelinecheck=false,
    skip=3pt
}
\caption{Hyperparameter sensitivity on equivalent-gate identification.
Bold parameter values denote the default settings, while bold metric values
indicate the best result within each parameter group.}
\label{tab:parameter-sensitivity}

\scriptsize
\renewcommand{\arraystretch}{0.95}

\begin{tabular*}{\columnwidth}{
@{\extracolsep{\fill}}
c c c c c c
@{}
}
\toprule
Parameter & Value & MAE $\downarrow$ & MSE $\downarrow$ & AP $\uparrow$ & F1 $\uparrow$ \\
\midrule

\multirow{4}{*}{$\alpha$}
& 0
& 0.0457
& 0.0132
& 0.0412
& \textbf{0.1235} \\

& \textbf{0.05}
& \textbf{0.0433}
& \textbf{0.0118}
& \textbf{0.0469}
& 0.1149 \\

& 0.1
& 0.0467
& 0.0132
& 0.0453
& 0.1079 \\

& 0.2
& 0.0437
& 0.0122
& 0.0429
& 0.1126 \\

\midrule

\multirow{6}{*}{$\lambda_{\mathrm{logic}}$}
& 0
& 0.0495
& 0.0135
& 0.0351
& 0.1007 \\

& 0.02
& 0.0448
& 0.0125
& 0.0439
& \textbf{0.1239} \\

& \textbf{0.05}
& \textbf{0.0433}
& \textbf{0.0118}
& \textbf{0.0469}
& 0.1149 \\

& 0.1
& 0.0462
& 0.0130
& 0.0440
& 0.1078 \\

& 0.2
& 0.0487
& 0.0140
& 0.0380
& 0.0995 \\

& 0.5
& 0.0695
& 0.0182
& 0.0383
& 0.0655 \\

\midrule

\multirow{6}{*}{$\lambda_{\mathrm{AND}}$}
& 0
& 0.0444
& 0.0126
& 0.0416
& 0.1076 \\

& \textbf{0.25}
& \textbf{0.0433}
& \textbf{0.0118}
& 0.0469
& 0.1149 \\

& 0.5
& 0.0453
& 0.0128
& 0.0413
& 0.1053 \\

& 1.0
& 0.0438
& 0.0120
& 0.0515
& \textbf{0.1353} \\

& 2.0
& 0.0447
& 0.0126
& \textbf{0.0620}
& 0.1251 \\

& 4.0
& 0.0451
& 0.0126
& 0.0424
& 0.1025 \\

\midrule

\multirow{4}{*}{$d_s$}
& 16
& 0.0456
& 0.0127
& 0.0470
& 0.1117 \\

& 32
& 0.0458
& 0.0130
& 0.0427
& 0.1056 \\

& \textbf{64}
& \textbf{0.0433}
& 0.0118
& 0.0469
& \textbf{0.1149} \\

& 128
& 0.0447
& \textbf{0.0115}
& \textbf{0.0507}
& 0.1128 \\

\midrule

\multirow{6}{*}{$L$}
& 3
& 0.0684
& 0.0198
& 0.0338
& 0.0917 \\

& 5
& 0.0541
& 0.0153
& 0.0380
& 0.1057 \\

& 7
& 0.0496
& 0.0141
& 0.0451
& 0.1112 \\

& \textbf{9}
& \textbf{0.0433}
& \textbf{0.0118}
& 0.0469
& \textbf{0.1149} \\

& 11
& 0.0445
& 0.0126
& \textbf{0.0501}
& 0.1040 \\

& 13
& 0.0454
& 0.0129
& 0.0406
& 0.1026 \\

\midrule

\multirow{3}{*}{Reconv.}
& None
& 0.0561 
& 0.0175
& 0.0229
& 0.0670  \\

& Flag
& 0.0454
& 0.0129
& 0.0403
& 0.1058 \\

& \textbf{Merge}
& \textbf{0.0433}
& \textbf{0.0118}
& \textbf{0.0469}
& \textbf{0.1149} \\

\bottomrule
\end{tabular*}

\end{table}

\subsection{Propagation and Architecture Details}
\label{app:architecture-details}

\paragraph{Initialization.}
Under the default configuration, each node is represented by an
8-dimensional input vector, and the hidden representation has dimension
$256$, with $d_h=128$ for each latent state.

The first propagation layer initializes the two latent states as
\begin{equation}
\mathbf{h}_{v,q}^{(1)}
=
\tanh\!\left(
\mathbf{W}_{q}^{(1)}
\left[
\operatorname{Agg}^{(0)}
\big(
\{\mathbf{x}_u\mid u\in\mathcal{N}_q(v)\}
\big)
\mathbin{\Vert}
\mathbf{x}_v
\right]
\right),
\qquad q\in\{0,1\},
\label{eq:initial-propagation}
\end{equation}
where $\operatorname{Agg}^{(0)}(\cdot)$ denotes the initialization
aggregation operator.

\paragraph{Subsequent Propagation.}
After initialization, the two latent states are concatenated into
$\mathbf{h}_v^{(l)}\in\mathbb{R}^{2d_h}$.
The encoded PI-dependency representation is introduced after the first layer
and retained throughout subsequent propagation, while the reconvergence
descriptor controls its contribution during node updates.

\paragraph{Network Configuration.}
The detailed architecture of the auxiliary modules is summarized in
Table~\ref{tab:architecture-details}. All three MLPs contain three linear
layers, with Batch Normalization, ReLU activation, and dropout after the first
two layers.

\begin{table}[h]
\centering
\caption{Architecture details of CircuitGate.}
\label{tab:architecture-details}
\scriptsize
\renewcommand{\arraystretch}{1.0}

\begin{tabular*}{\linewidth}{@{\extracolsep{\fill}}lcccc@{}}
\toprule
Module
& Input
& Hidden
& Output
& Dropout \\
\midrule

Support $\mathrm{MLP}_q$
& $2d_s+4=132$
& 128
& 128
& 0.1 \\

$\mathrm{MLP}_s$
& $4d_h=512$
& 128
& 128
& 0.1 \\

$\mathrm{MLP}_g$
& $6+d_h=134$
& 128
& 128
& 0.1 \\

Probability head
& 256
& 256
& 1
& 0.2 \\

\bottomrule
\end{tabular*}
\end{table}

\paragraph{Node Handling.}
The processed graphs contain three gate types: PI, AND, and explicit NOT
nodes. PI nodes are identified by the PI gate type and initialize their
positive-parity support with their graph-local PI index. Primary outputs are
identified by zero fanout and receive an explicit output-status feature.
Reconvergence descriptors are constructed only for two-input AND nodes;
all other nodes use a zero reconvergence descriptor. All node types otherwise
share the same propagation and update rules.

\end{document}